\documentclass[sn-mathphys]{sn-jnl}% Math and Physical Sciences Reference Style
\usepackage{graphicx}
\usepackage{amsmath}
\usepackage{subfigure}
\usepackage{caption2}
\usepackage{diagbox}
\usepackage{algorithm}
\usepackage{algpseudocode}

\jyear{2021}%

\theoremstyle{thmstyleone}%
\theoremstyle{thmstyletwo}%

\theoremstyle{thmstylethree}%

\begin{document}

\title[A-CVAE]{Advanced Conditional Variational Autoencoders (A-CVAE): Towards interpreting open-domain conversation generation via disentangling latent feature representation}


\author[]{\fnm{Ye} \sur{Wang}}
\author[]{\fnm{Jingbo} \sur{Liao}}
\author*[]{\fnm{Hong} \sur{Yu*}} \email{yuhong@cqupt.edu.cn}
\author[]{\fnm{Guoyin} \sur{Wang}}
\author*[]{\fnm{Xiaoxia} \sur{Zhang}}
\author[]{\fnm{Li} \sur{Liu}}

\affil[]{\orgdiv{Chongqing Key Laboratory of Computational Intelligence}, \orgname{Chongqing University of Posts and Telecommunications}, \orgaddress{\city{Chongqing}, \postcode{400065}, \country{China}}}


\abstract{Currently end-to-end deep learning based open-domain dialogue systems remain black box models, making it easy to generate irrelevant contents with data-driven models. Specifically, latent variables are highly entangled with different semantics in the latent space due to the lack of priori knowledge to guide the training. To address this problem, this paper proposes to harness the generative model with a priori knowledge through a cognitive approach involving mesoscopic scale feature disentanglement. Particularly, the model integrates the macro-level guided-category knowledge and micro-level open-domain dialogue data for the training, leveraging the priori knowledge into the latent space, which enables the model to disentangle the latent variables within the mesoscopic scale. Besides, we propose a new metric for open-domain dialogues, which can objectively evaluate the interpretability of the latent space distribution. Finally, we validate our model on different datasets and experimentally demonstrate that our model is able to generate higher quality and more interpretable dialogues than other models.}




\maketitle

\section{Introduction}\label{sec1}
Human-machine dialogue systems have been developed for almost 70 years, from the introduction of the Turing Test \cite{moor2003turing} to the present day. Over the years it has acquired increasing attention for its enormous commercial and research value in areas such as semantic assistants and chatbots. Although the black-box deep learning model of human-machine dialogue is continuously progressing regarding the conversation quality, the end-to-end model still lacks interpretability, which indicates that the model is procedurally unverifiable and incomprehensible. Therefore, the range of applications is comparably limited especially when particularly high-demanding applications require manifestly decision-making logic. Currently, several deep learning models \cite{ref2}\cite{refQin}\cite{refLiu}\cite{refGangadharaiah}\cite{refSaha} have been proposed to build dialogue systems, mainly divided into task-oriented and non-task-oriented dialogue application systems. Task-oriented dialogue systems are designed to accomplish specific tasks in a domain. Regarding the non-task-oriented system, it is principally an open-domain dialogue system for entertainment, which aims to generate relevant responses. In open domain dialogue systems, the traditional seq2seq generation model \cite{ref4} encodes the dialogue into a fixed vector of knowledge representations, and inputting the same questions will generate the same responses. In contrast, Transformer-based dialogue generation models \cite{ref5} abandon the basic paradigm of using a circular recursive structure to encode sequences and use a self-attention mechanism to compute the hidden state of a sequence. Although self-attention mechanism can better model dependencies in long sequences, Transformer still does not address the problem that seq2seq cannot generate diverse text. The adversarial generative nets (GAN) \cite{refGAN} achieves Nash equilibrium by playing a game between two networks with opposite goals, the discriminative network and the generative network. The discriminative network continuously improves the level of discrimination, while the generative network is able to generate more realistic sentences. By borrowing the modeling form of Conditional GAN \cite{refCGAN}, generative adversarial nets can be extended to applications of dialogue generation. Currently GANs are already capable of generating good quality text without pre-training. Variational Autoencoders (VAEs) \cite{ref6} encodes text sequences as probability distributions in the latent space instead of deterministic vectors, which can better model text diversity and achieve category controllability of generation. However, in practical text generation tasks, data often appear in pairs, such as questions and responses in dialogue systems. Conditional Variable Autoencoders (CVAE) \cite{ref19} adds a conditional encoder to VAE to encode ground-truth responses and some conditions in dialogues, making it more suitable for text generation tasks. Specifically, the latent variables of deep models potentially contain richer semantic features, attracting the researchers to explore the relationship between the feature and the generated data. Therefore, deep latent variable models have additionally become an important work in the study of dialogue systems. To summarise our contributions:
\begin{itemize}

\item To solve the semantic entanglement problem in open-domain dialogue system for generating relevant conversations, we integrate the macro-level guided-category knowledge and micro-level open-domain dialogue data for the training, which enables the model to disentangle the latent variables within the mesoscopic scale.

\item Because hidden features in the black box dialogue models are hard to measure, we propose a new metric to objectively evaluate the interpretability of potential features. Extensive experiments are also designed to analyze the interpretability of our model using this metric.

\item To validate the model in terms of interpretability, diversity and quality of text generation, adequate experiments have been performed on different datasets and comprehensive analyses have been provided.
\end{itemize}

\section{Related Works}\label{sec2}
Our work is closely related to the study of deep latent variable models. Kingma et al. \cite{ref6} proposed a variational autoencoders and applied it to the image domain. Later, Bowman et al. \cite{ref7} applied VAEs to the field of natural language processing. Zhao et al. \cite{ref8} proposed two models based on VAEs, DI-VAE and DI-VST. The models discover interpretable semantics through automatic encoding or contextual prediction, thus solving the problem that traditional dialogue models can not output interpretable actions. Serban et al. \cite{refSerban} proposed a noval latent variable neural network structure, VHRED, with latent stochastic variables that span a variable number of time steps. And the model is suitable for text generation conditional on long contexts. Zhao et al. \cite{ref9} proposed a dialogue generation model based on conditional variational autoencoders and incorporated external knowledge to improve the interpretability of the model. Gao et al. \cite{ref10} introduced a discrete variable with explicit semantics in CVAE, using the semantic distance between latent variables to maintain good diversity among the sampled latent variables. Wang et al. \cite{refSCVAE} proposed the semantic-aware conditional variant autoencoder (S-CVAE) model. S-CVAE can generate diverse conversational text by utilizing embedded classifiers and feature decoupling modules. Shi et al. \cite{ref11} introduce an exponential mixture distribution to replace the Gaussian prior and capture the hidden semantic relations between the mixture components and the data, resulting in a more interpretable latent space. Besides, they also decompose the evidence lower bound and derive the loss term that causes mode collapse, which solves the Mode Collapse problem to some extent. Pang et al. \cite{ref12} proposed a text generation model based on the energy of the latent space. They coupled continuous latent variables and discrete categories in a priori distributions to guide the generator to generate high-quality, diverse, and interpretable texts.\\
Through a series of studies, we found that in the latent space of a common deep latent variable model, dialogues with different semantics are entangled together, resulting in many different semantic data distributed around a dialogue with a definite semantics. Due to the randomness of sampling, it is likely to sample latent variables with wrong semantics. Then the quality of the generated model is improved if the latent variables with accurate semantics can be sampled in the latent space. We differ from these studies in that we train different categories of dialogues separately by a dialogue generation task, and then use the approximate posterior distribution of the deep latent variable model as the prior knowledge of the corresponding category. Finally, we use the prior knowledge to guide the latent space decoupling, so that the dialogues of different categories are distributed in the corresponding regions of the latent space. After disentangling, the model is able to better sample the latent variables of the exact semantic categories, thus generating high-quality, interpretable, and diverse texts.

\section{Proposed Scheme}\label{sec3}
\subsection{Variational Autoencoders}
\label{subsec:pagestyle}
VAEs \cite{ref6} are able to encode texts as probability distributions in latent space instead of deterministic vectors, and sampling from latent space can obtain rich latent variables, which can model the diversity of texts. Suppose X is the observed variable and z is the latent variable. First, VAE samples the latent variable z from the prior distribution $\mathbf{p}(\boldsymbol{z})$ of the latent variable, and then generates the observed variable z according to the conditional distribution $\mathbf{P}(\mathbf{X} \mid \mathbf{z})$. And the probability distribution of the observed variable is decomposed as shown in Eq.1.
\begin{align}
P(X)=\int_{z} p(z) P(X \mid z) d z
\end{align}
Eq.1 has an integral over the latent variable, and the latent variable z is continuous. If the integration is calculated directly, all latent variables need to be sampled, so it is difficult to calculate and optimize directly. To solve the problem, VAEs introduce the variational probability distribution $\mathbf{q}_{\phi}(\mathbf{z} \mid \mathbf{X})$ to approximate the true posterior distribution $\mathbf{p}(\mathbf{z} \mid \mathbf{X})$ of the latent variable, and approximate $\log P(X)$ as the  evidence lower bound, as shown in Eq.2.
\begin{align}
\log P(X) \geq E_{q_{\phi}(z \mid X)}\left[\log P_{\theta}(X \mid z)\right]-K L\left(q_{\phi}(z \mid X) \mid p(z)\right)
\end{align}

\begin{figure}[htpb]%
\centering
\includegraphics[width=0.9\textwidth]{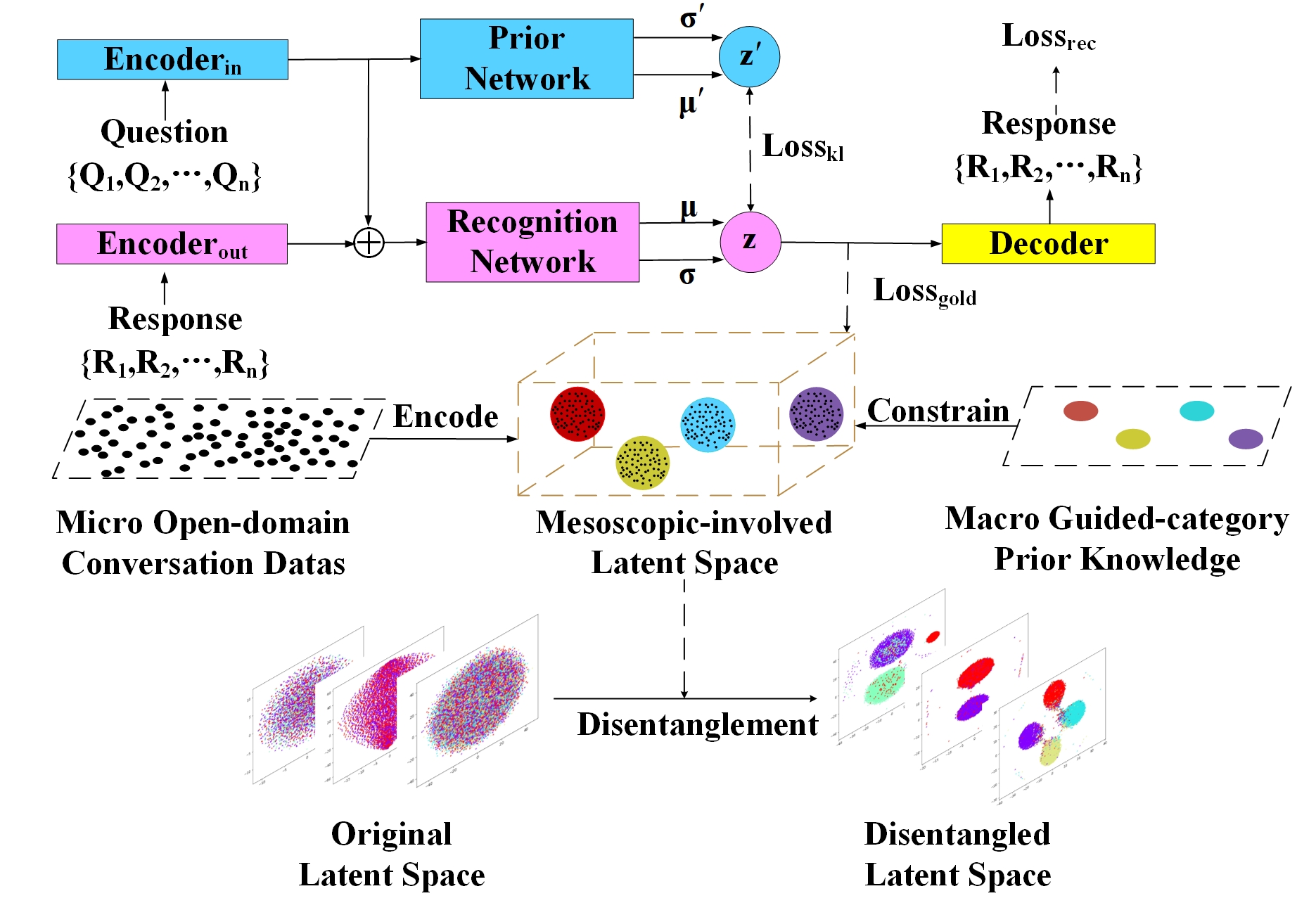}
\caption{A-CVAE model diagram.}\label{fig1}
\end{figure}

\subsection{Disentangled Conditional Variational Autoencoders}
\label{subsec:pagestyle}
Although the traditional deep latent variable model introduces latent variables to optimize the model, the whole model is still a black box model. So we do not know what the latent variables represent. Deep latent variable models are more interpretable if we can specify what the latent variables represent. The latent variable is represented as a 200-dimensional multidimensional Gaussian distribution with each dimension assumed to be independent from each other. We can minimize the KL dispersion value between the latent variable distribution and the knowledge distribution to approximate both distributions, making the latent variable distribution learn to external knowledge. Our scheme is proposed with the name of Disentangled Conditional Variational Autoencoders (A-CVAE) (Fig. \ref{fig1}). A-CVAE can disentangle the jumbled latent space in the deep latent variable model by introducing external knowledge, so that different categories of dialogues are encoded into different regions of the latent space, and thus the semantically accurate latent variables can be sampled during random sampling. Taking the external knowledge as emotion as an example, by disentangling the latent space, we can not only know the sampled latent variable is related to emotion, but also know the emotion is positive or negative. So the process of sampling latent variable from disentangled latent space enhances the interpretability of dialogue text generation. To solve the problem of how to introduce external knowledge so that the latent space can be separated by category, we will use the CVAE model to train the category prior knowledge so that it is represented as a Gaussian distribution. And then category prior knowledge will be used to disentangle the latent space.\\
Our approach is divided into two steps, representing prior knowledge and training the model using prior knowledge.\\
\subsubsection{Module Definition.}
The input in the dialogue task is divided into a question sequence $\left\{\mathbf{C}_{1}, \mathbf{C}_{2}, \cdots, \mathbf{C}_{n}\right\}$ denoted as C and a response sequence $\left\{\mathbf{X}_{1}, \mathbf{X}_{2}, \cdots, \mathbf{X}_{n}\right\}$ denoted as X. To apply the text to the deep model, we need to embed words into the word vector space. $e(W)$ denotes embedding the word W into the word space, and W can be either C or X. For training, we use questions and responses to sample latent variables from an approximate posterior distribution to generate dialogue text. For testing, we use only the questions to sample the latent variables from the prior distribution, thus generating the text. Our model consists of five main modules, the input encoder $\operatorname{Enc}_{i n}(\cdot)$, the output encoder $\operatorname{Enc}_{out}(\cdot)$, the recognition network $\operatorname{RecogNetq}_{\phi(z \mid X, C)}$, the prior network $\text { PriorNetp }_{\varphi(z \mid C)}$ and the decoder $\text { Dec }(\cdot)$. The input encoder is used to encode the questions in the conversation, the output encoder is used to encode the responses in the conversation, the recognition network is responsible for generating the mean and variance of the approximate posterior distribution, the prior network is responsible for generating the mean and variance of the prior distribution, and the decoder is used to decode the latent variables and generate the responses.
\\
\subsubsection{The priori knowledge representation. } To fuse the category knowledge into the model and then disentangle the latent space, we need to train the CVAE on the dialogue task and  use the latent space distribution of the CVAE as the category prior knowledge representation. First, the dataset is divided into k sub-datasets by category. Then data are randomly selected from each sub-dataset and trained on CVAE to obtain a priori knowledge representations $\mathcal{N}\left(\mu_{k}, \sigma_{k}^{2} I\right)$ for each category. k denotes the kth category. And the prior knowledge representation of the category is the latent space distribution of CVAE, which is a multidimensional Gaussian distribution. We name the prior knowledge representation as gold Gaussian. The gold Gaussian is used to guide the training of the model, which is approximated by the Kullback-Leibler divergence (KL) to the multidimensional Gaussian distribution in the latent space of the model during training.\\
\subsubsection{Advanced Conditional Variational Autoencoders.} In our A-CVAE model, we propose the optimized objective function by extra adding our proposed optimized loss function for bootstrapping the latent space distribution to the standard loss of the CVAE. In CVAE, our goal is to fit the conditional probability distribution $P(X \mid C)$. We calculate $P(X\mid C)$ by introducing the latent variable z, as shown in Eq.3.
\begin{align}
P(X \mid C) & = \int_{z} p(z \mid C) P(X \mid z, C) d z
\end{align}
Since there is an integral over the continuous variable z in the middle of Eq.3, and the integral is difficult to be calculated. So an approximate solution is obtained by means of variational extrapolation, and the solution is called the evidence lower bound(ELBO), as shown in Eq.4.
\begin{equation}
\begin{aligned}
\mathrm{ELBO} = E_{q_{\phi}(z \mid X, C)}\left[\log P_{\theta}(X \mid z, C)\right]-K L\left(q_{\phi}(z \mid X, C) \| p_{\varphi}(z \mid C)\right)
\end{aligned}
\end{equation}
$E_{q_{\phi}(z \mid X, C)}\left[\log P_{\theta}(X \mid z, C)\right]$ is the loss function of the reconstructed response text. We use the input encoder $\operatorname{Enc}_{i n}(\cdot)$  to encode the question sequence to obtain the feature $r_{C}$ of the question C, and we use the output encoder $\operatorname{Enc}_{out}(\cdot)$ to encode the response to obtain the feature $r_{X}$ of the response sequence X. And We use $r_{C}$ and $r_{
X}$ as inputs to the recognition network $\operatorname{RecogNetq}_{\phi(z \mid X, C)}$ to obtain the mean $\mu$ and variance $\sigma^{2}$ of the approximate posterior distribution. Also, $r_{C}$ is used as the input of the prior network $\text { PriorNetp }_{\varphi(z \mid C)}$ to calculate the mean $\mu^{\prime}$ and variance ${\sigma^{\prime}}^{2}$ of the prior distribution. $q_{\phi}(z \mid X, C)$ is the approximate posterior distribution, which is used during training. $P_{\theta}(X \mid z, C)$ is a conditional probability distribution for generating the response text using the latent variable z and the input condition C, which is fitted by the decoder $\text { Dec }(\cdot)$. Also the processes sampling the latent variables are not derivable, and to be able to optimize the model by the gradient descent algorithm, we use the reparameterization trick to make all the processes derivable, as shown in Eq.5.
\begin{align}
z & = \mu+\sigma * \varepsilon, \qquad\varepsilon \sim \mathcal {N}(0, I)
\end{align}
$KL\left(q_{\phi}(z \mid X, C) \| p_{\varphi}(z \mid C)\right)$ is the KL dispersion of the prior distributions $p_{\varphi}(z \mid C)$ and approximate posterior distributions $q_{\phi}(z \mid X, C)$. Due to the specificity of the conversation generation task data, during testing we have only question sequences as input and no reference response sequences. Therefore, in the training and testing phases, we sample in two different multidimensional Gaussian distributions, respectively, the approximate posterior distribution $\mathcal{N}\left(\mu, \quad \sigma^{2} I\right)$ and the prior distribution $\mathcal{N}\left(\mu^{\prime}, \quad {\sigma^{\prime}}^{2}I\right)$. To make the two distributions closer, the algorithm uses the KL distance to approximate them. The parameters $\mu$, $\sigma^{2}$, $\mu^{\prime}$, ${\sigma^{\prime}}^{2}$ are fitted by the prior network and the recognition network, respectively. Where $\phi$, $\varphi$, $\theta$ are the parameters of two Gaussian distributions.\\
Our proposed optimization loss is shown in Eq.6. This loss function can guide the KL distance between the approximate posterior distribution and the gold Gaussian distribution of each class to present variability in different dimensions. $\mathcal{N}\left(\mu_{k}, \sigma_{k}^{2} I\right)$ is the gold Gaussian distribution obtained by pre-training the kth category. During training, the KL value is calculated by selecting the corresponding gold Gaussian distribution according to the category of the dialogue text, thus approximating the posterior distribution and the Gaussian distribution to optimize the parameters.
\begin{align}
\label{gold}
\mathcal{L}_{\text {gold }} & = K L\left(\mathcal{N}\left(\mu_{k}, \sigma_{k}^{2} I\right) \| q_{\phi}(z \mid X, C)\right)
\end{align}
The total loss during training of the A-CVAE model is the reconstructed expectation loss plus the KL scatter of the prior and approximate posterior distributions, plus our proposed optimisation loss. $\beta$, $\lambda$ are the kl term coefficient to avoid the KL posteriori collapse problem. As shown in Eq.7:
\begin{equation}
\begin{aligned}
\mathcal{L} & = -E_{q_{\phi}(z\mid X, C)}\left[\log P_{\theta}(X \mid z, C)\right]\\&+\beta K L\left(q_{\phi}(z \mid X, C) \| p_{\varphi}(z \mid C)\right) \\
&+\lambda K L\left(\mathcal{N}\left(\mu_{k}, \sigma_{k}^{2} I\right) \| q_{\phi}(z \mid X, C)\right)
\end{aligned}
\end{equation}

\begin{algorithm}
\caption{A-CVAE training procedure}\label{algo1}
\begin{algorithmic}[1]
\Require Data samples (C, X) = ${(C_{i}, X_{i})}_{i=1}^{N}$.
\Ensure Parameter($\phi$, $\varphi$, $\theta$) convergence.
\State Initialize the parameters ($\phi$, $\varphi$, $\theta$).
\If{Train}
        \State \textbf{for} iter from 1 to max\_iter \textbf{do}
        \State \hspace*{0.15in} Sampling $(C, X)$ from the training set.
        \State \hspace*{0.15in} Calculate question text features $r_{C}$$\leftarrow$$\operatorname{Enc}_{i n}(\cdot)(e(C))$.
         \State \hspace*{0.15in} Calculate response text features $r_{X}$$\leftarrow$$\operatorname{Enc}_{out}(\cdot)(e(X))$.
        \State \hspace*{0.15in} Calculate  $\mu$, $\sigma$$\leftarrow$$RecogNetq_{\phi(z \mid X, C)}(r_{X},r_{C})$.
        \State \hspace*{0.15in} Calculate $\mu^{\prime}$, $ \sigma^{\prime}$$\leftarrow$$PriorNetp_{\varphi(z \mid C)}(r_{C})$.
        \State \hspace*{0.15in} Reparameterize $z = \mu+\sigma * \varepsilon$.
        \State \hspace*{0.15in} Calculate generated text $\bar{X}$$\leftarrow$$\text { Dec }(z)$.
        \State \hspace*{0.15in} Calculate gold KL loss by Eq.6.
        \State \hspace*{0.15in} Calculate variational evidence lower bound by Eq.4.
        \State \hspace*{0.15in} Update $\phi$, $\varphi$ and $\theta$ by gradient ascent by Eq.7.
        \State \textbf{end for}
\EndIf
\end{algorithmic}
\end{algorithm}

\subsection{Interpretability Evaluation Index for Dialog}
\label{subsec:pagestyle}
We propose a metric named Interpretability Evaluation Index for Dialog (IEID) for objectively evaluating the interpretability of latent space distribution. Since we sample the latent space to get latent variables containing category semantics, the text generated by the decoder using latent variables will contain category information. Therefore, we train a text classifier using CVAE with reference responses from the dataset as input to the input encoder $Encoder_{in}$. Besides, the category labels is the input of the output encoder $Encoder_{out}$, and the decoder reconstructs the category labels. Since the responses are generated with latent variables containing category information, the semantics of the output contains related category information. Under this assumption, we further implement the specific classifiers to classify the generated responses and reference output, aiming to verify the correctness of the proposed hypothesis. Therefore, we define the proportion of reference responses and generated responses that are classified into the same category as an evaluation metric to evaluate the interpretability of the latent space distribution. Here below is the description of IEID: $t$ stands for ground-truth response and $g$ represents generated response. $Clf(\cdot)$ denotes the classifier. $Label_{t}$ defines the classifier takes $t$ as the classification result of the input. $Label_{g}$ denotes the classifier takes $g$ as the classification result of the input and $n$ is the number of conversations in the test set.
\begin{align}
label_{t} = C l f(t),\qquad label_{g} = C l f(g)
\end{align}

\begin{align}
IEID= \frac{\sum_{i = 1}^{n} f\left(label_{t_{i}}, label_{g_{i}}\right)}{n}
\end{align}

\begin{align}
f(label_{t_{i}},label_{g_{i}}) = \left\{\begin{array}{ll}
1 & label_{t_{i}} = label_{g_{i}} \\
0 & label_{t_{i}} \neq label_{g_{i}}
\end{array}\right.
\end{align}

\section{Experiments}
\label{sec:typestyle}
\subsection{Dataset}
\textbf{DD dataset:} DailyDialog Dataset \cite{ref13} is a multi-round conversation dataset for everyday chat scenarios, which has less noise than previous corpora. DD dataset covers several major topics of life, and is annotated with the action and emotion of each conversation. In our experiments, we used the emotion and action labels from the dataset. Due to the uneven distribution of conversations across categories in the dataset, we aggregated the conversations into three categories (no emotion, negative, positive) according to the emotion category. We also decomposed the multi-round conversations into single-round conversations.\\
\textbf{ED dataset:} EmpatheticDialogues dataset\cite{EDdataset} is a large-scale multi-round conversation dataset collected on Amazon Mechanical Turk, containing 24,850 one-to-one open domain conversations. The dataset provides labels for emotions in 32 categories. We aggregated the conversations by emotion category into positive and negative.\\
For the labels in the dataset, we have two main uses: dividing the sub-dataset according to the labels and selecting the golden Gaussian to optimize the model according to the labels of the conversation during model training.

\subsection{Automatic Evaluation Metrics}
We use three evaluation metrics to assess the quality of text generation, an evaluation metric to evaluate the diversity of the generated texts, and our proposed metric (IEID) to evaluate the interpretability of the hidden space distribution.
\par\textbf{BLEU} \cite{ref14}  is a fast, inexpensive and language-independent method for assessing the quality of text generation. This method can replace review by skilled humans when rapid or frequent evaluation is required. And BLEU calculates the co-occurrence word frequency for both sentences. 
\par\textbf{ROUGE} \cite{ref15} is proposed to estimate text quality by calculating co-occurrence frequencies. Unlike BLEU the words in ROUGE can be discontinuous. \par\textbf{METEOR} \cite{ref16} is made more relevant to manual discrimination by calculating the relationships between synonyms, roots and affixes. 
\par\textbf{Distinct-2} \cite{ref17} is used to evaluate the diversity of text generated by the generative model. A higher value indicates that the generated text is richer, while a lower value indicates that the generated text contains a large amount of repetitive, generic and meaningless text.\\

\subsection{Benchmark Methods}
We used the same dataset on the dialogue generation task and compared it with the model below. 
\par \textbf{Seq2seq} \cite{ref4}, breaking through the traditional fixed-size input problem framework, converts an input sequence of indefinite length into an output sequence of indefinite length. It also incorporates a Global Attention mechanism \cite{ref18}\cite{refAbmrnn} to notice more textual information; 
\par \textbf{Transformer} \cite{ref5}, eschewing the traditional recurrent neural network, and using the self-attention mechanism to encode and decode the text. The self-attention mechanism also allows the Transformer to encode longer sequences of text; 
\par \textbf{Vanilla VAE} \cite{ref6}, encoding text into a continuous latent space rather than a fixed vector. Sampling the latent variables from a Gaussian distribution increases the diversity of the generated text;  
\par \textbf{Vanilla CVAE} \cite{ref19}, which adds a conditional encoder to VAE, is able to encode the response text, sentiment and other conditions into the model. Such a structure is more suitable for dialogue generation tasks; 
\par \textbf{CVAE with VADE} \cite{ref20}, introducing a Gaussian mixture model in the latent space, to obtain a better representation of the text through the clustering task.
\begin{figure}[htb]%
\centering
 
\subfigure[Original latent space distribute]{
    \begin{minipage}[t]{0.50\linewidth}
        \centering
        \includegraphics[width=1.651in]{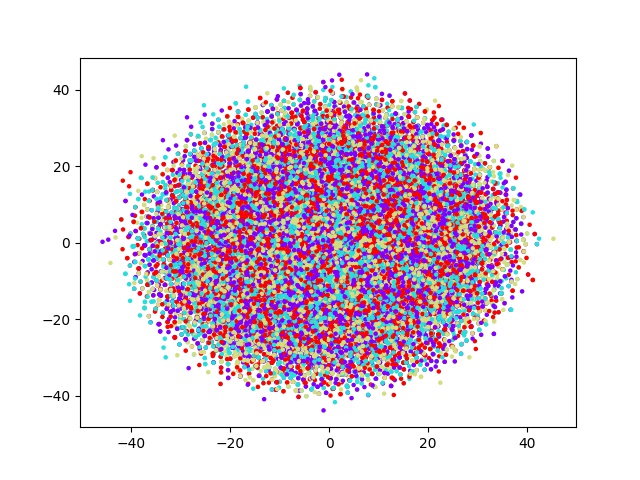}\\
        \vspace{0.02cm}
        \includegraphics[width=1.651in]{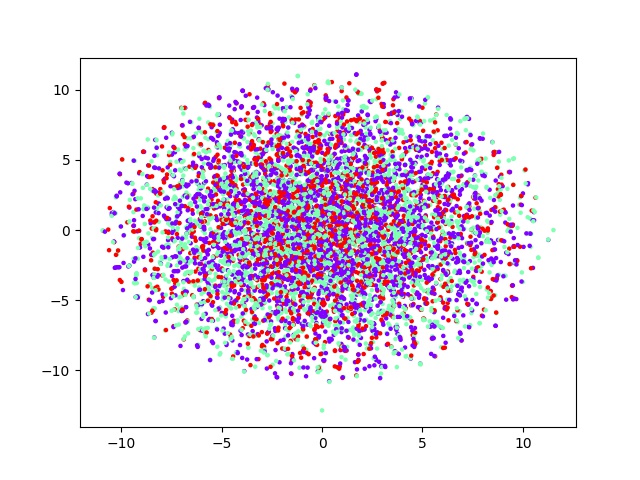}\\
        \vspace{0.02cm}
        \includegraphics[width=1.651in]{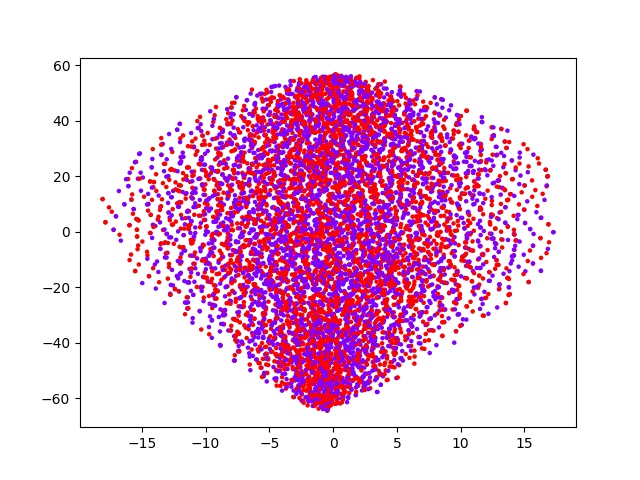}\\
        \vspace{0.02cm}
        %\caption{fig1}
    \end{minipage}%
}%
\subfigure[Disentangled latent space distribute]{
    \begin{minipage}[t]{0.50\linewidth}
        \centering
        \includegraphics[width=1.651in]{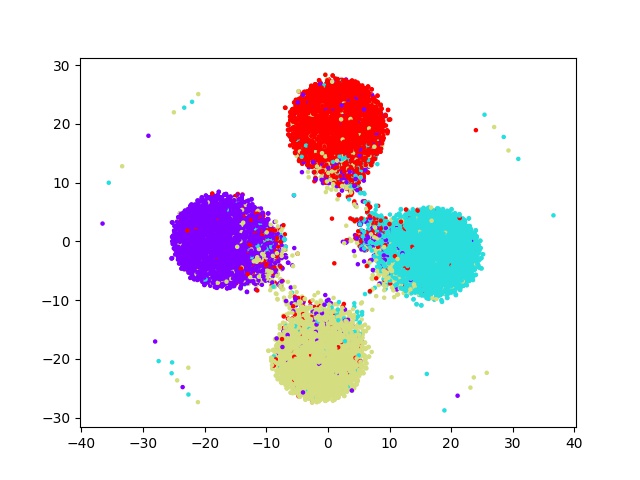}\\
        \vspace{0.02cm}
        \includegraphics[width=1.651in]{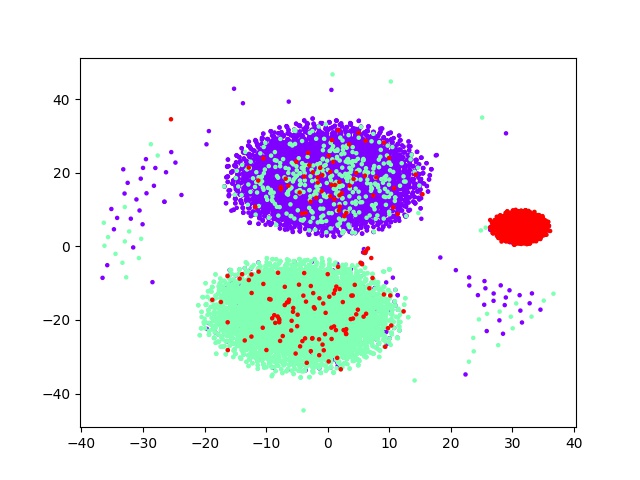}\\
        \vspace{0.02cm}
        \includegraphics[width=1.651in]{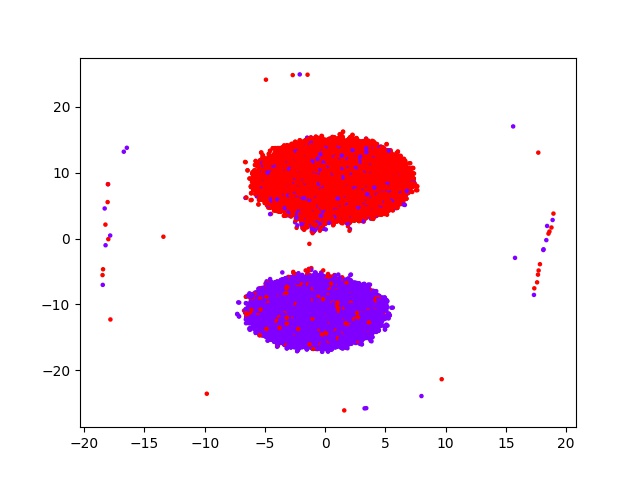}\\
        \vspace{0.02cm}
        %\caption{fig1}
    \end{minipage}%
}%
 
\caption{Visualization results of the latent space distribution for different datasets. The datasets from top to bottom are DailyDialog dataset with action label, DailyDialog dataset with emotion label, and EmpatheticDialogues dataset with emotion label. }
\label{fig2}
\end{figure}
\subsection{Implementation Details}
The words in the dataset were processed to produce a word list dictionary containing 6918 words. The dictionary contains four special tokens, $\langle PAD \rangle$, $\langle SOS \rangle$, $\langle EOS \rangle$ and $\langle UNK \rangle$. $\langle PAD \rangle$ denotes padding values, $\langle SOS \rangle$ denotes the start of sentence, $\langle EOS \rangle$ denotes the end of sentence and $\langle UNK \rangle$ denotes words that do not appear in the word list. The word embedding vector is 300 dimensions. The encoder and decoder of our proposed model utilize LSTM \cite{ref21} as the basic unit, with hidden size 300, latent variable 200 dimensions, and periodic annealing using the KL annealing algorithm to avoid KL posterior collapse. The recognition and prior networks are multilayer perceptrons with the number of hidden layer neurons set to 250. In the training phase, the model uses the Adam optimizer. The initial learning rate is 0.0001, the learning rate decays by 0.01 after each epoch, and the coefficient of the kl term increases to 1 after 10,000 updates.

\subsection{Results and analysis}
\label{subsec:pagestyle}
\subsubsection{Disentangled result.} We downscaled the original latent space distribution of the deep latent variable model and the disentangled latent space distribution into two dimensions by t-SNE and visualized them. The original latent space represents the latent space without disentangling, and the disentangled latent space represents the A-CVAE disentangled latent space. Fig. \ref{fig2} shows the visualization results. We can see that the original latent space is a miscellaneous distribution, which cannot distinguish between different classes of data in the latent space. The deep latent variable generation model obtains the latent variables by random sampling in the latent space, and it is very difficult to accurately sample the corresponding semantic latent variables in the miscellaneous latent space. Due to the randomness of sampling and the chaotic data distribution of each category in the original latent space, it is easy for the model to sample other semantic latent variables in the mixed latent space. After disentangling the latent space by our method, the data of the same category are closer to each other in the latent space, while the data of different categories are further away from each other. We can find that after disentangling, the same class of data is distributed around one same category data in the latent space. And although the sampling process is still random, we can still sample the corresponding class of data more accurately in the disentangled latent space. The comparison experiments in the next subsection also show that extracting semantically accurate latent variables in the disentangled latent space can improve the quality, interpretability, and diversity of the generated text.

\begin{table*}[htb]\centering
\caption{The automatic evaluation results of all compared methods on the DD dataset using action labels.}\label{tab1}
\begin{tabular}{c|c|c|c|c}
\hline Method              & BLEU                & METEOR              & ROUGE               & dist-2        \\
\hline Seq2seq      & 10.14±0.08 & 10.52±0.09 & 30.15±0.11 & 0.17 \\
Transformer & 10.11±0.08 & 11.16±0.09 & \textbf{33.09±0.11} & 0.25 \\
Vanilla VAE         & 10.85±0.10 & 11.52±0.11 & 30.73±0.11 & 0.14 \\
Vanilla CVAE                & 11.50±0.11          & 11.95±0.12          & 31.12±0.11          & 0.14\\
CVAE with VADE                                        & 11.49$\pm$0.20           & 12.23$\pm$0.23          & 32.44$\pm$0.21                   & 0.22 \\
\hline A-CVAE                                        & \textbf{12.28$\pm$0.23} & \textbf{12.98$\pm$0.26} & 32.36$\pm$0.22 & \textbf{0.26}\\
\hline
\end{tabular}
\end{table*}

\begin{table*}[htb]\centering
\caption{The automatic evaluation results of all compared methods on the DD dataset using emotion labels.}\label{tab2}
\begin{tabular}{c|c|c|c|c}
\hline Method              & BLEU                & METEOR              & ROUGE               & dist-2        \\
\hline Seq2seq      & 11.15±0.22 & 11.89±0.26 & 30.99±0.25 & 0.30  \\
Transformer & 10.44±0.19 & 11.03±0.21 & \textbf{31.59±0.25} & \textbf{0.42} \\
Vanilla VAE         & 10.01±0.17 & 10.48±0.19 & 28.09±0.21 & 0.18 \\
Vanilla CVAE &11.17$\pm$0.22 &11.50$\pm$0.25 &29.46$\pm$0.23 &0.25                                         \\
CVAE   with VADE & 10.12$\pm$0.19          & 10.50$\pm$0.21          & 28.07$\pm$0.22                & 0.23                                           \\
\hline A-CVAE            & \textbf{12.59$\pm$0.29} & \textbf{13.22$\pm$0.33} & 30.72$\pm$0.28                & 0.28                         \\\hline 
\end{tabular}
\end{table*}

\begin{table*}[htb]\centering
\caption{The automatic evaluation results of all compared methods on the  Empathetic Dialogu dataset using emotion labels.}\label{tab3}
\begin{tabular}{c|c|c|c|c}
\hline Method              & BLEU                & METEOR              & ROUGE             & dist-2             \\
\hline 
Seq2seq                      & 6.86$\pm$0.08                 & 8.04$\pm$0.10                   & 31.03$\pm$0.12                & 0.21                               \\
Transformer                  & 6.40$\pm$0.08                 & 7.90$\pm$0.09                   & 31.69$\pm$0.13       & \textbf{0.41}                         \\
Vailla   VAE                 & 7.40$\pm$0.09                 & 8.37$\pm$0.10                   & 30.37$\pm$0.12                & 0.19                                         \\
Vanilla   CVAE               & 7.37$\pm$0.09                 & 8.39$\pm$0.10                   & 30.79$\pm$0.12                & 0.22                                          \\
CVAE   with VADE             & 6.94$\pm$0.09                 & 8.09$\pm$0.10                   & 30.17$\pm$0.12                & 0.14                                           \\\hline 
A-CVAE               & \textbf{7.75$\pm$0.10}        & \textbf{8.56$\pm$0.11}          & \textbf{31.72$\pm$0.13}       & 0.23                               \\\hline  
\end{tabular}
\end{table*}

\begin{table*}[htb]\centering
\caption{The result of IEID. }\label{tab4}%
\begin{tabular}{c|c|c|c}
\hline
\diagbox{Method}{Dataset} &Action Of DD  & Emotion of DD & Emotion of ED \\
\hline Seq2seq   & 0.19 & 0.56 & 0.62 \\
Transformer      & 0.24 & \textbf{0.61} & \textbf{0.72} \\
Vanilla VAE      & 0.21 & 0.52 & 0.63 \\
Vanilla CVAE     & 0.44 &0.57 &0.63 \\
CVAE with VADE   & 0.39 &0.56 &0.63\\
\hline A-CVAE    & \textbf{0.44} &0.56 &0.65\\
\hline
\end{tabular}
\end{table*}

\begin{figure}[htb]%
\centering
 
\subfigure[Error bar result of action]{
    \begin{minipage}[t]{0.50\linewidth}
        \centering
        \includegraphics[width=0.9\textwidth]{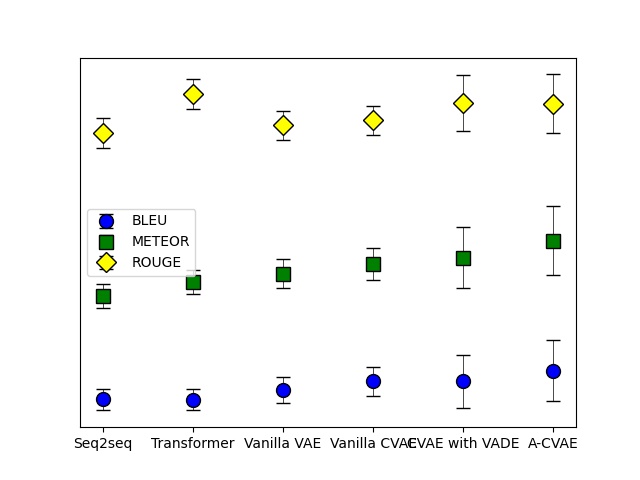}\\
        \vspace{0.02cm}
        %\caption{fig1}
    \end{minipage}%
}%
\subfigure[Error bar of emotion]{
    \begin{minipage}[t]{0.50\linewidth}
        \centering
        \includegraphics[width=0.9\textwidth]{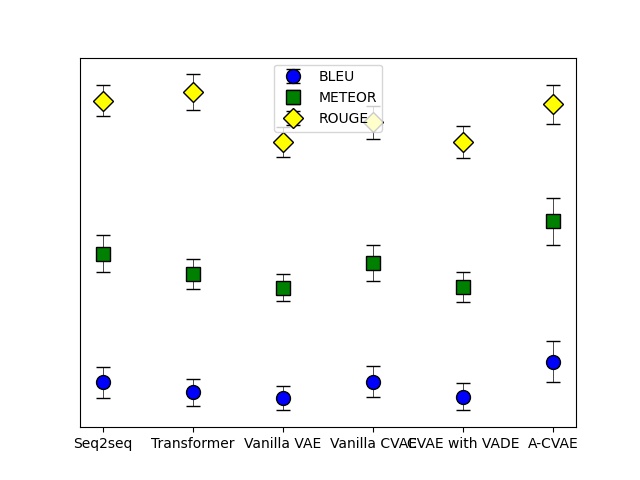}\\
        \vspace{0.02cm}
        %\caption{fig1}
    \end{minipage}%
}%
 
\caption{Error bar results for metrics with different generation models on DailyDialog dataset. }
\label{fig3}
\end{figure}

\begin{figure}[htb]
\begin{minipage}{0.48\linewidth}
\centering
\includegraphics[width=0.9\textwidth]{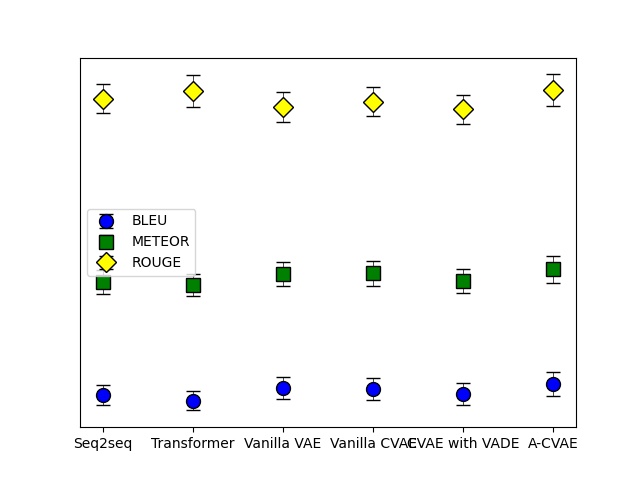}
\caption{Error bar results for metrics with different generation models on EmpatheticDialogues dataset.}\label{fig4}
\end{minipage}\quad
\begin{minipage}{0.48\linewidth}  
\centering
\includegraphics[width=0.9\textwidth]{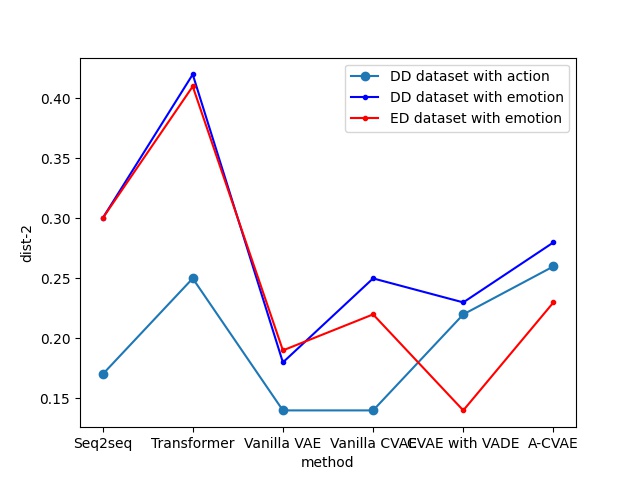}
\caption{Diversity assessment metric dist-2 line chart of different models under different datasets.}\label{fig5}
\end{minipage}
\end{figure}

\subsubsection{Automatic evaluation result.} Table \ref{tab1}, Table \ref{tab2} and Table \ref{tab3} show the results of the evaluation metrics for our model and the comparison models on different datasets. Table \ref{tab1} shows the results on the DailyDialog dataset with action labels, Table \ref{tab2} shows the results on the DailyDialog dataset with emotion labels, and Table \ref{tab3} shows the results on the EmpatheticDialogues dataset with emotion labels.\\
The tables show that our model outperforms most comparative models on several metrics that assess the quality of text generation. This indicates that our model is able to generate high quality, diverse conversations. In particular, our model outperforms all comparative models on the BLEU and ROUGE metrics, and is only lower than Transformer on ROUGE. With the fusion of category information, we can sample higher quality latent variables from the disentangled latent space and thus generate higher quality text. And the results illustrates that external prior knowledge has an impact on text generation, and that sampling the correct category of latent variables enhances text generation. Our model achieves excellent results with different datasets and different labels, indicating that our model has good generalization ability.\\
Fig. \ref{fig3} and Fig. \ref{fig4} show the error bar results of different models for different metrics on different datasets. We can find that although our model achieves good results on several different evaluation metrics, the value of error bar is relatively large, indicating that the text generated by our model is not very stable and sometimes the quality of the generated text is poor. In our future work, we will study how to generate more stable and high-quality texts.\\
Fig.\ref{fig5} shows the results of the diversity evaluation index in the form of a line graph. From the line graph, we can find that our model is better than most the other comparison models except Transformer. We have analyzed the generated text and found that Transformer tends to generate some longer dialogues, which is due to Transformer's self-attention mechanism can encode longer sequences of text, so the generated text is also longer than the normal model. And the metric dist-2 is used by stitching all generated texts together and then taking n words as a group, and finally calculating the proportion of non-repeated n-gram phrases to the total n-gram phrases, so the longer texts may get higher dist-2 scores.\\

\subsubsection{IEID result.} Interpretability Evaluation Index for Dialog (IEID) is the metric we proposed in section 3.2 for evaluating the interpretability of the distribution of the latent space. Table \ref{tab4} shows the IEID results for different models on different datasets. In the DailyDialog dataset with action tag our model is higher than most of the comparison models, indicating that the disentangled latent space has a positive impact on text generation. While on the DailyDialog dataset with emotion labels and EmpatheticDialogues dataset our model does not differ much from the other models, after analysis, we found that it is due to the low accuracy of the classifier used in the process of computing the metrics. Moreover, the emotion tag only has three categories of no emotion, positive and negative on the DailyDialog dataset, and only two categories of positive and negative on the EmpatheticDialogues dataset, so the generated text The probability of the generated text and the reference text being classified into the same category becomes larger, resulting in insignificant results. However, it can still be seen that our model has better interpretability than the common generative model.

\begin{table*}[htb]\centering
\caption{The automatic evaluation results of multi-dimensional latent space on the DD dataset using action and emotion labels. Ratio represents the ratio of the dimensions of common features, action, and emotion in the latent space.}\label{tab5}
\begin{tabular}{c|c|c|c|c}
\hline Ratio    & BLEU                & METEOR              & ROUGE               & dist-2        \\
\hline50:50:100 & 12.30±0.23          & 13.14±0.26          & 32.38±0.22          & \textbf{0.25} \\
50:50:100(Math) & 12.15±0.22          & 12.97±0.25          & 32.37±0.21          & 0.24          \\
\hline50:75:75  & 11.83±0.22          & 12.54±0.25          & 31.81±0.21          & 0.24          \\
50:75:75(Math)  & 12.03±0.22          & 12.76±0.25          & 31.96±0.22          & 0.24          \\
\hline50:100:50 & \textbf{12.53±0.23} & \textbf{13.36±0.26} & \textbf{33.08±0.22} & 0.24          \\
50:100:50(Math) & 11.96±0.22          & 12.77±0.25          & 32.30±0.21          & 0.24 \\\hline        
\end{tabular}
\end{table*}
\subsubsection{Other analysis.}

In our previous experiments, we were using each tag separately. So that the latent space is disentangled according to the action tag or the emotion tag. In this subsection, we use both action tags and emotion tags to disentangle the latent space. Meanwhile, we abstracted a common feature which represents the part that is not disentangled according to the tag. So the latent space is decomposed into three parts, the common feature which is not disentangled , the action feature which is disentangled by action tags, and the emotion feature which is disentangled by emotion tags. And we construct three different hidden spaces with the three parts in different proportions. In addition, when describing the motivation of our method, it is mentioned that different gold Gaussians are different between different dimensions, so the disentangling can be done by approximating the latent space to the corresponding gold Gaussians through KL distance, and finally our experiments also prove this point. And the gold Gaussian is pre-trained by following different classes of data, which is essentially a multi-dimensional Gaussian distribution with several different parameters. So it is natural to think that the gold Gaussian may not necessarily need to be pre-trained to get it. So we can also construct our own gold Gaussian with several different parameters through mathematical methods. We take the expectation of the Gaussian distribution 
changing in steps of 1 from 0 (including 0) to the left and right and the variance changing in steps of 1 from 0 (not including 0) to the left and right. Thus, several Gaussian distributions with different parameters are constructed as gold Gaussians. Table 5 shows the results of the experiments, where the ratio is the proportion of three different features, the labeled math represents our own constructed gold Gaussian, and the unlabeled represents the gold Gaussian trained by a single class of dataset. From the results, we can see that both the pre-trained gold Gaussian and our own constructed gold Gaussian can get better results than the baseline, which shows that the gold Gaussian can be constructed by mathematical methods without the need to get it by training. Meanwhile, we can see from the results that after dividing the latent space into 3 different parts, we can still get better results than baseline, but the results are slightly worse than A-CVAE with a single label. The main reason for this is that the data set is not balanced in terms of categories, and it is not possible to balance the action and emotion at the same time, so that the labels of both categories are equally distributed in the data set.

\section{Conclusion}
\label{sec:pagestyle}
In this paper, for further disentangle the latent space, we propose the A-CVAE model, which integrates macroscopic category prior knowledge and micro dialogue data to guide the latent feature disentangling into a meso-scale. Specifically, the correct-semantic latent variables are sampled from the  disentangled latent space, which enhances the interpretability of the model in the latent space and generate related responses. Additionally, we propose a new metric for objectively evaluating the interpretability of the latent space distribution in open-domain dialogues. The experimental results show that on the DD dataset and ED dataset, our model generates higher quality and more diverse dialogue text, providing extra interpretable latent spaces than other models.

\section{Acknowledgements}

This work was partly supported by National Key R\&D Program of China (2021YFF0704100), the National Natural Science Foundation of China (62136002, 61936001 and 61876027), the Science and Technology Research Program of Chongqing Municipal Education Commission (KJQN202100627 and KJQN202100629), and the National Natural Science Foundation of Chongqing (cstc2019jcyj-cxttX0002,cstc2022ycjh-bgzxm0004), respectively.

\bibliography{sn-bibliography}% common bib file


\end{document}